\documentclass[10pt,journal,compsoc]{IEEEtran}

\ifCLASSOPTIONcompsoc
  \usepackage[nocompress]{cite}
\else
  \usepackage{cite}
\fi
\usepackage[T1]{fontenc}
\usepackage{graphicx}
\usepackage{amsmath}
\usepackage{url}
\usepackage{amssymb}
\usepackage{fontawesome5}
\usepackage{booktabs}
\usepackage{algorithmic}
\usepackage{algorithm}
\usepackage{tikz}
\usepackage{marvosym}
\usepackage{etoolbox}
\usepackage{ragged2e}
\usepackage{enumitem}
\usepackage{caption}
\usepackage{multirow}
\newcommand*{\circled}[1]{\lower.7ex\hbox{\tikz\draw (0pt, 0pt)%
    circle (.5em) node {\makebox[1em][c]{\small #1}};}}
\robustify{\circled}
\usepackage[colorlinks,linkcolor=blue]{hyperref}

\ifCLASSINFOpdf
\else
\fi
\begin{document}
\title{EfficientSync: Real-Time Lip Synchronization via Deformation-Based Reference Texture Mixing}

\author{Fa-Ting~Hong$^*$,~Runzhen Liu$^*$,~Luchuan Song,~Hongmin Cai,~and~Chuhua~Xian\textsuperscript{\Letter}
        % <-this % stops a space
\IEEEcompsocitemizethanks{\IEEEcompsocthanksitem * Equal Contribution.
\protect 
\IEEEcompsocthanksitem \textsuperscript{\Letter} Corresponding author.
\protect 
\IEEEcompsocthanksitem Fa-Ting Hong is with the Department
of Computer Science and Engineering, The Hong Kong University of Science and Technology, Hong Kong SAR.
\protect
E-mail: fatinghong@gmail.com.
\IEEEcompsocthanksitem Runzhen Liu, Hongmin Cai and Chuhua Xian are with South China University of Technology. E-mail: csalunaticat@mail.scut.edu.cn, hmcai@scut.edu.cn, chhxian@scut.edu.cn.
\IEEEcompsocthanksitem Luchuan Song is with the Computer Science Department at the University of Rochester. E-mail: slc0826@mail.ustc.edu.cn.
}% <-this % stops an unwanted space
% \thanks{Manuscript received April 19, 2005; revised August 26, 2015.}
}


% \markboth{SUBMISSION TO IEEE TRANSACTIONS ON PATTERN ANALYSIS AND MACHINE INTELLIGENCE}%
% {Shell \MakeLowercase{\textit{et al.}}: Bare Demo of IEEEtran.cls for Computer Society Journals}
\markboth{Journal of \LaTeX\ Class Files,~Vol.~14, No.~8, August~2015}%
{Shell \MakeLowercase{\textit{et al.}}: Bare Demo of IEEEtran.cls for Computer Society Journals}


\IEEEtitleabstractindextext{%
\begin{abstract}
Audio-driven lip synchronization aims to manipulate the mouth region of a talking-face video so that it conforms to the driving audio, while preserving head pose, identity, and background. By its nature, the task entails local editing of the mouth rather than global resynthesis of the face. Nevertheless, prevailing approaches reconstruct the entire lower face via heavy GAN- or diffusion-based decoders. This design incurs substantial inference latency. More critically, it hallucinates intra-oral high-frequency details such as teeth and lip wrinkles instead of preserving the subject's authentic textures. We contend that the principal bottleneck in identity preservation lies not in the scarcity of reference frames, but in the absence of a mechanism that faithfully transfers the genuine textures already present in them. Motivated by this insight, we present EfficientSync, a real-time deformation-based framework that explicitly retains reference textures rather than resynthesizing them. First, we reformulate multi-reference fusion as a channel-wise selection problem, in place of channel-coupled convolutions or texture-disrupting spatial self-attention. To this end, we propose the Dynamic Texture Mixer. It evaluates each spatially aligned reference in a global context and aggregates them via channel-wise weighted summation. This preserves textural integrity at a substantially reduced computational cost. Second, we introduce the Spatio-Temporal Shifted Adaptive Masking strategy to reconcile the long-standing conflict between mask coverage and background fidelity. It decomposes the source frame into lip-generation conditions and an independent background prior. This suppresses lower-face leakage and allows the synthesized mouth to blend seamlessly into the original background. Third, we devise STAR Sampling, a zero-overhead pre-processing procedure that retrieves the sharpest and topologically most diverse frames, since deformation quality is ultimately bounded by the reference pool. Extensive experiments on the HDTF and VFHQ datasets demonstrate that EfficientSync attains state-of-the-art visual quality and identity preservation while operating at 166 FPS on a single GPU. Video demos are available at \url{https://alunaticat.github.io/EfficientSync/index.html}.
\end{abstract}

}

\maketitle

\IEEEdisplaynontitleabstractindextext

% \fi
\IEEEpeerreviewmaketitle

\IEEEraisesectionheading{\section{Introduction}\label{sec:intro}}
\begin{figure}[t]
  \centering
    \includegraphics[width=\linewidth]{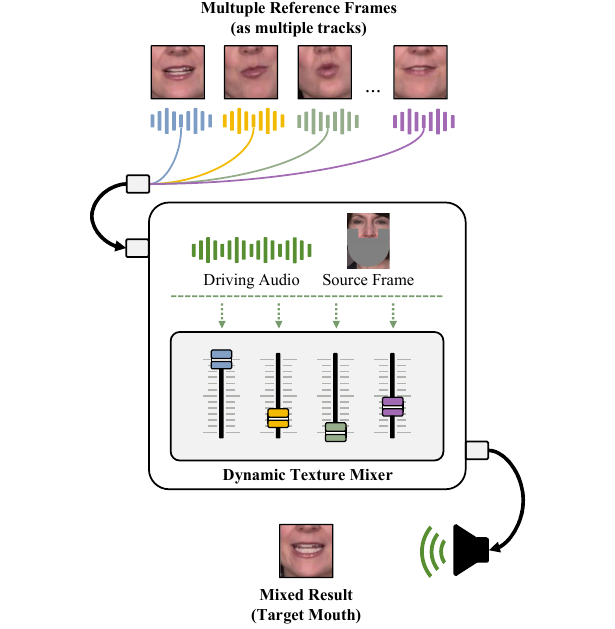}
    \caption{\textbf{Motivation of the Dynamic Texture Mixer.} Rather than reconstructing the mouth from scratch, we treat each reference as a track on a mixing console and adjust its volume, that is, its aggregation score, according to the driving audio and the source identity.}
  \label{figs:dtm_motivation}
\end{figure}
\IEEEPARstart{L}{ip} synchronization (lip-sync) aims to manipulate the mouth movements of a person in a video to match a given driving audio, while preserving the remaining attributes such as head pose, identity, and background. This task underpins a broad spectrum of real-world applications, including film dubbing and post-production, multilingual video translation, virtual avatars and digital humans, video conferencing, and accessibility tools for the hearing impaired. By its nature, lip-sync modifies only the mouth region of the source frames and leaves all other facial attributes intact, so it should in principle be both computationally efficient and faithful to the source identity. In practice, however, existing methods continue to struggle on both fronts, which severely constrains their deployment in latency-sensitive and quality-critical scenarios.

Because lip-sync operates on videos rather than a single image, one might expect the abundant reference frames of the same subject to already supply its mouth-region identity, leaving little room for the identity loss that plagues one-shot generation. Yet the mere availability of references does not guarantee that this identity information is faithfully exploited, and the failure manifests in two distinct stages. The first is the way the lower face is handled. Most methods~\cite{prajwal2020lip,zhang2023dinet,li2024latentsync} apply a single fixed mask that erases the entire lower half and compel a heavy generative or diffusion decoder to redraw it from scratch, even though large portions of this region, such as the chin, jawline, and surrounding skin, require no regeneration at all. Worse, the reference frames are rarely aligned with the target in head pose or mouth openness, so the decoder must resolve appearance and geometry simultaneously, a process in which fine details such as tooth edges and lip wrinkles are easily blurred. The second is the way references are fused. Reference features are repeatedly merged with frame-varying audio features through stacked convolutions and attention layers, where the time-varying audio signal tends to dominate the time-invariant identity cues, so the authentic textures carried by the references are gradually smoothed away. When the target phoneme demands a mouth shape that no reference clearly exhibits, the network, lacking a faithful exemplar, falls back to fabricating the intra-oral appearance from audio alone, which is precisely where blurred teeth and implausible textures emerge. Taken together, these observations reveal that the difficulty lies not in the scarcity of references, but in two coupled deficiencies: the absence of a mechanism that genuinely retains and faithfully transfers the real textures already present in those references, and the absence of a principled way to edit only the mouth while leaving the surrounding regions truly untouched.

This bottleneck recurs across the main families of lip-sync methods. GAN-based pixel synthesis, represented by Wav2Lip~\cite{prajwal2020lip} and its follow-ups~\cite{yin2022styleheat,karras2019style,liang2022expressive}, concatenates references with the masked source and trains a generator under a sync discriminator, but suffers from low resolution and blurred intra-oral regions. Diffusion-based generation, including DiffTalk~\cite{shen2023difftalk}, Diff2Lip~\cite{mukhopadhyay2024diff2lip}, and LatentSync~\cite{li2024latentsync}, synthesizes the mouth in a latent space and injects identity through reference embeddings, yet incurs the high latency of multi-step sampling and still recovers identity through a generative prior rather than from real reference pixels. Person-specific modeling, such as RAD-NeRF~\cite{tang2025real} and ER-NeRF~\cite{li2023efficient}, preserves identity by overfitting a dedicated model per speaker, at the cost of failing to generalize to unseen identities. The family closest in spirit to ours is warping-based synthesis, exemplified by MCNet~\cite{hong2023implicit} and StyleSync~\cite{guan2023stylesync}, which deforms a reference toward the target pose with an estimated flow field. However, it typically warps only a single reference and then refines the warped result with a decoder that resynthesizes the texture, which partially negates the very benefit of operating on real reference pixels.

% \begin{figure}[t]
%   \centering
%   \includegraphics[width=0.9\linewidth]{figs/radar.pdf}
%   \caption{\textbf{Visualization of Quantitative Evaluation.} Comparison of six lip-sync models in the cross-audio setting on seven key metrics. Results are normalized, with the best-performing model scaled to the outer edge, and the worst towards the center.}
%   \label{figs:radar}
% \end{figure}

Motivated by these observations, we advocate a deformation-based paradigm that preserves reference textures instead of hallucinating them. Rather than regenerating the mouth, we predict an independent flow field for each of a diverse set of references and warp them toward the target topological shape, which explicitly retains pre-existing high-frequency details such as teeth and lip wrinkles rather than destructively resynthesizing them. Because these flow fields are computed in parallel, the approach fully exploits GPU parallelism, so that incorporating multiple references does not compromise inference speed. Building on this principle, we develop EfficientSync. Turning the paradigm into a working system, however, raises two questions, which we address in turn.

\emph{How should the spatially aligned references be fused without destroying their textures?} Conventional convolutional fusion concatenates references along the channel dimension and processes them with stacked convolutions, which couples channels and averages out fine details; spatial self-attention, the fusion operator favored by diffusion backbones, instead fractures continuous textures into patches and scales quadratically with spatial resolution. We observe that, once spatially aligned, the references form a pool of candidate features sharing identical semantic channels at the same locations, so fusion reduces to a channel-wise selection problem: for each channel, choosing the most suitable candidate under the current context. We therefore propose the \textbf{Dynamic Texture Mixer (DTM)}, which, as illustrated in Fig.~\ref{figs:dtm_motivation}, treats each reference as a track on a mixing console and aggregates them by context-aware, channel-wise weighted summation, preserving texture integrity at a markedly lower computational cost. Because the quality of this aggregation is ultimately bounded by the reference pool, yet random sampling often returns blurry or near-duplicate frames, we further introduce \textbf{STAR Sampling}, a zero-overhead pre-processing step that retrieves the sharpest and topologically most diverse frames to strengthen the candidate pool without adding inference latency.

\begin{figure}[ht!]
  \centering
  \includegraphics[width=0.8\linewidth]{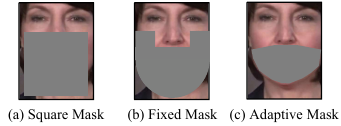}
  \caption{\textbf{Existing mask strategies.} (a) Square mask used in DINet~\cite{zhang2023dinet}. (b) Fixed mask used in LatentSync~\cite{li2024latentsync}. (c) Adaptive mask used in IPTalker~\cite{liu2026identity}.}
  \label{figs:mask}
\end{figure}

\emph{How should the synthesized mouth be composited back into the original background?} This is where the second deficiency surfaces: keeping the surrounding regions truly untouched requires local editing that generates only the mouth and fuses it with the source. Yet the prevailing fusion paradigm faces a long-standing dilemma over the mask range (Fig.~\ref{figs:mask}): a fixed large mask blocks lower-face leakage but covers part of the background and thus leaves conspicuous seam artifacts during fusion, whereas an adaptive mask preserves the background but induces the Shortcut Learning problem, where the model infers the lip shape from the surrounding chin rather than from the audio and consequently produces inaccurate articulation under cross-audio inference. We attribute this dilemma to treating the source background as a fixed, immutable environment, and instead decouple the source frame into global attributes that condition lip generation and a background prior that is independent of the lip shape. Realizing this idea, we propose the \textbf{Spatio-Temporal Shifted Adaptive Masking (STAM)} strategy, which conditions lip generation on a fixed large mask to fully suppress leakage while supplying background textures through an independent background reference, thereby blending the synthesized mouth seamlessly into the original background and eliminating boundary artifacts.

Integrating these designs, EfficientSync delivers state-of-the-art high-fidelity synthesis and identity preservation while operating at a real-time speed of 166 FPS. Our primary contributions are summarized as follows:
\begin{itemize}
    \item We advocate a deformation-based paradigm that retains real reference textures rather than hallucinating them, and propose the Dynamic Texture Mixer, which reformulates multi-reference fusion as a channel-wise selection problem and aggregates aligned references under global context with low computational overhead.
    \item We propose the Spatio-Temporal Shifted Adaptive Masking strategy, which decouples the source frame into lip-generation conditions and an independent background prior, resolving the conflict between mask range and background fidelity.
    \item We introduce STAR Sampling, which improves the quality of the reference pool, and hence the final generation, while incurring no additional inference cost.
    \item Extensive experiments demonstrate that EfficientSync attains state-of-the-art visual quality and identity preservation at a real-time speed of 166 FPS on a single GPU.
\end{itemize}

\section{Related Works}\label{sec:related}
\subsection{Audio-driven Portrait Animation}
Audio-driven portrait animation aims to synthesize talking head videos with natural head movements and facial expressions from a single static reference image and a driving audio signal. Early mainstream approaches primarily relied on explicit intermediate representations. For example, StyleTalk~\cite{ma2023styletalk} leverages 3D Morphable Models (3DMM)~\cite{2002A} to derive expression parameters for synthesizing stylized faces. MakeItTalk~\cite{zhou2020makelttalk} predicts a sequence of facial landmarks from speech content and speaker identity to drive the subsequent image rendering. Subsequently, NeRF-based~\cite{mildenhall2021nerf} methods introduced implicit neural representations to improve 3D consistency and view stability. As a representative, AD-NeRF~\cite{guo2021ad} constructs a dynamic Neural Radiance Field that is directly conditioned on audio features to output video sequences. DFRF~\cite{shen2022learning} emphasizes identity generalization, equipping the model with the ability to rapidly adapt to unseen facial appearances using only sparse observation frames. Recently, diffusion models have brought significant breakthroughs to this domain. EMO~\cite{tian2024emo} employs an end-to-end denoising architecture driven by weak conditional guidance to produce highly expressive facial dynamics. Hallo~\cite{xu2024hallo} introduces a reference net that shares the exact topology of the main diffusion network to extract texture information from the source image. AniPortrait~\cite{wei2024aniportrait} adopts a two-stage paradigm, first estimating intermediate landmarks from the inputs and then utilizing a diffusion model to translate these spatial guides into final video frames. 

While these methods offer flexible control over poses and expressions, they are sub-optimal for lip-sync tasks. Lip-sync requires preserving the original pose, lighting, and background. However, portrait animation models tend to hallucinate unintended global movements and viewpoint shifts, disrupting the spatial-temporal consistency of the source video.

\begin{figure*}[ht!]
    \centering
    \includegraphics[width=\textwidth]{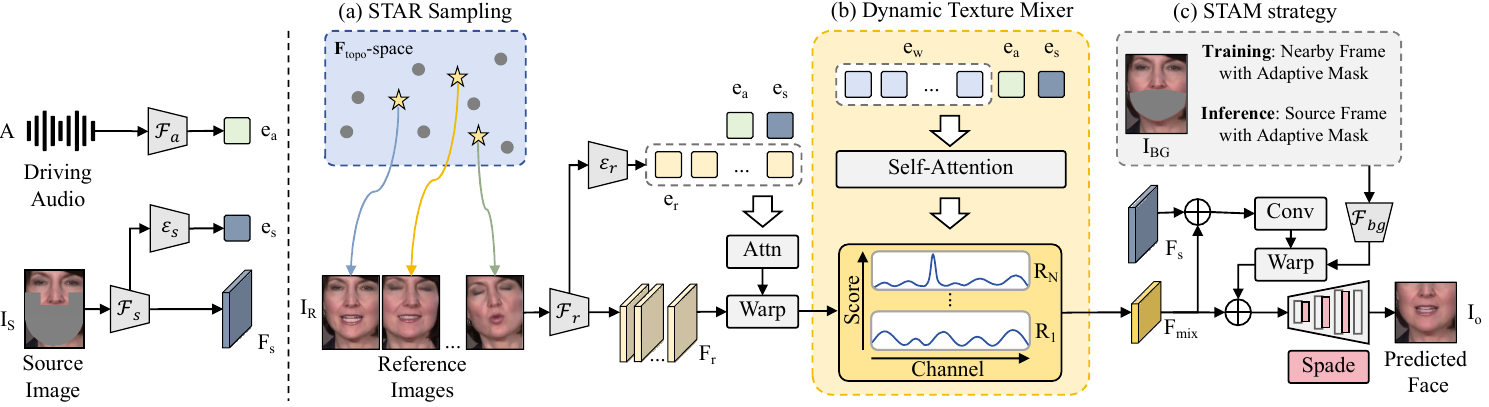}
    \caption{\textbf{The framework or our EfficientSync.} (a) STAR Sampling first selects, from the source video, a reference pool that is both sharp and topologically diverse, supplying high-quality material for deformation at no inference cost. (b) The Dynamic Texture Mixer then aggregates the spatially aligned references into a high-fidelity mouth feature through a context-aware, channel-wise selection mechanism. (c) The Spatio-Temporal Shifted Adaptive Masking strategy finally composites the synthesized mouth back into the source frame, blending it seamlessly with the preserved background.}
    \label{figs:framework}
\end{figure*}
\subsection{Audio-driven Lip Synchronization}
The audio-driven lip synchronization task focuses specifically on video dubbing, aiming to precisely replace the mouth region of the lower-half face based on target audio while keeping the background and upper-face pose of the source video unchanged. Early pioneering works established the masked inpainting technical paradigm.  Wav2Lip~\cite{prajwal2020lip} incorporated a pre-trained audio-visual synchronization discriminator (SyncNet) to compel the generative network to capture highly accurate lip-audio correlations. Following this foundation, researchers introduced multi-dimensional feature transformations and advanced spatial operations. DINet~\cite{zhang2023dinet} architects target mouth shapes by executing spatial deformations directly on the reference features. Through self-attention mechanisms, IPLAP~\cite{zhong2023identity} synthesizes talking heads by inferring lip and jaw landmarks conditioned on audio, reference points, and pose priors. VideoReTalking~\cite{cheng2022videoretalking} proposes a sequential pipeline that first edits the expressions of original frames, serving them as references to guide the audio-driven lip-sync generation via masked frame blending. With the evolution of generative models, diffusion-based architectures have significantly improved the visual realism of the synthesized frames. MuseTalk~\cite{zhang2024musetalk} executes cross-attention fusion between audio and image signals within a frozen VAE latent space. LatentSync~\cite{li2024latentsync} formulates a direct channel-wise concatenation of reference frames, masked inputs, and noisy latents before processing them through a U-Net. KeySync~\cite{bigata2025keysync} specifically targets boundary artifacts by designing an anti-leakage mask processing pipeline coupled with an interpolation generation strategy. 

However, existing methods often rely on heavy generative networks or multi-step denoising processes. This introduces high computational costs and hardware thresholds, resulting in slow inference speeds that hinder widespread deployment. Furthermore, while diffusion models excel at generating high-fidelity overall image quality, fine-grained deterministic control over mouth details is often lacking. This stochastic nature can easily lead to inconsistent intra-oral textures and a subsequent loss of the original identity.

\section{Methodology}\label{sec:methodology}
Given a source frame, a driving audio, and a video of the same subject, our goal is to synthesize a lip-synchronized face that follows the audio while preserving identity, pose, and background. 
As illustrated in Fig.~\ref{figs:framework}, the framework is organized around a deformation-based backbone and comprises three designs that directly address the questions raised above. (a) STAR Sampling first selects, from the source video, a reference pool that is both sharp and topologically diverse, supplying high-quality material for deformation at no inference cost. (b) The Dynamic Texture Mixer then aggregates the spatially aligned references into a high-fidelity mouth feature through a context-aware, channel-wise selection mechanism. (c) The Spatio-Temporal Shifted Adaptive Masking strategy finally composites the synthesized mouth back into the source frame, blending it seamlessly with the preserved background. 

\subsection{Overview}
We extract audio features $\mathbf{A}$ using Whisper~\cite{radford2023robust}. To prevent information leakage, a fixed mask is applied to the source frame to construct the masked source $\mathbf{I}_S$. To explicitly preserve background pixels, we additionally retain a background reference $\mathbf{I}_{BG}$ obtained via an adaptive mask conceptually similar to IPTalker~\cite{liu2026identity}. The $N$ reference frames $\{\mathbf{I}_R^i\}_{i=1}^N$ are not drawn at random but selected by STAR Sampling, described next. All source and reference frames are uniformly cropped to a $416 \times 320$ face region for strict spatial alignment.

As illustrated in Fig.~\ref{figs:framework}, our framework follows a deformation-based paradigm and proceeds in three stages. First, STAR Sampling (Fig.~\ref{figs:framework}(a)) retrieves a sharp and topologically diverse reference pool from the source video, from which lightweight extractors produce the source feature $\mathbf{F}_s$ and the reference features $\{\mathbf{F}_r^i\}_{i=1}^N$. Since the references differ from the target in head pose and mouth openness, each is warped to the target pose before fusion: an attention block first derives a per-reference warping condition from the source feature, the audio, and the reference features, and a flow field predicted under this condition is then applied by the differentiable warping operation of IPTalker~\cite{liu2026identity}, yielding the aligned features $\{\mathbf{F}_w^i\}_{i=1}^N$. Second, the Dynamic Texture Mixer (Fig.~\ref{figs:framework}(b)) fuses these aligned references into a high-fidelity mouth feature $\mathbf{F}_{mix}$ through a context-aware, channel-wise selection mechanism that preserves the reference textures intact. Finally, the Spatio-Temporal Shifted Adaptive Masking strategy (Fig.~\ref{figs:framework}(c)) composites $\mathbf{F}_{mix}$ back into the source frame, reusing the same warping step to align the background reference before a SPADE decoder renders the synchronized output $\mathbf{I}_o$. The following subsections detail our three core designs: STAR Sampling, the Dynamic Texture Mixer, and STAM.
\begin{figure}[t]
    \centering
    \includegraphics[width=0.45\textwidth]{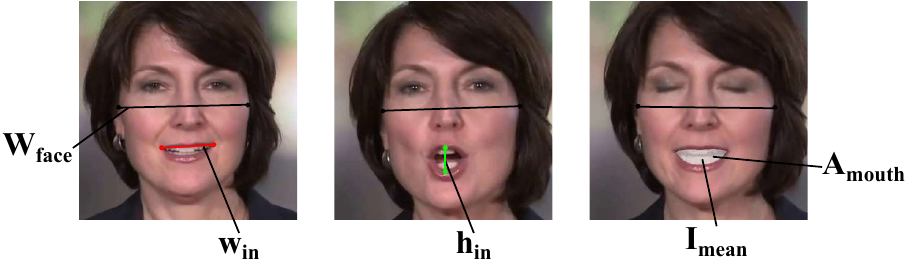}
    \caption{\textbf{Foundational metrics for STAR Sampling.} Visualization of the key geometric and textural measurements extracted from facial landmarks.}
    \label{figs:topo}
\end{figure}

\begin{figure*}[ht!]
    \centering
    \includegraphics[width=\textwidth]{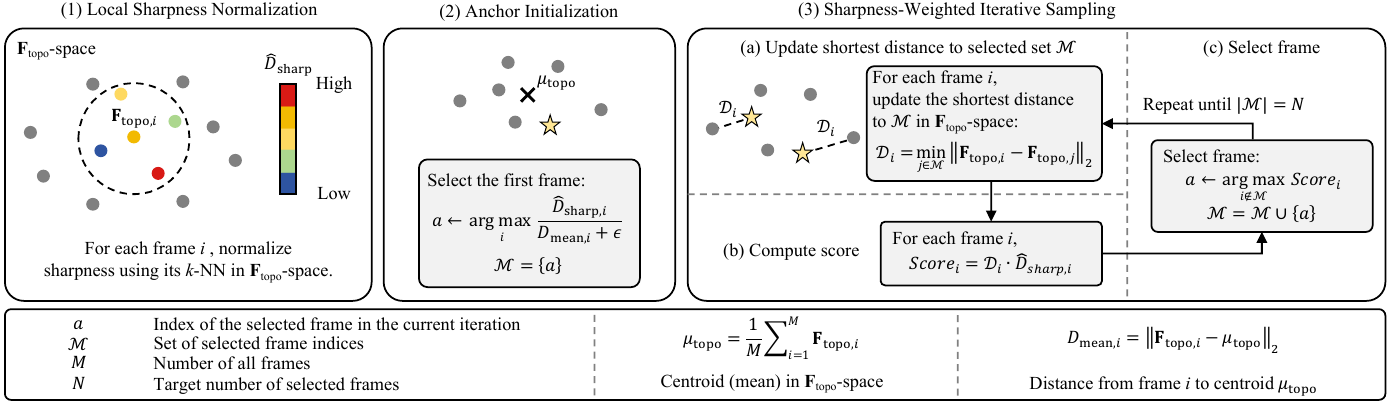}
    \caption{\textbf{Illustration of STAR Sampling.} (1) Each frame's sharpness is normalized within its $k$-nearest neighbors in $\mathbf{F}_\text{topo}$-space to obtain a relative clarity $\hat{D}_{\text{sharp}}$. (2) The first anchor maximizes clarity over centrality, $\frac{\hat{D}_{\text{sharp},i}}{D_{\text{mean},i} + \epsilon}$. (3) Each subsequent anchor maximizes the clarity-weighted farthest-point distance until $N$ references are selected.}
    \label{figs:star}
\end{figure*}
\subsection{STAR Sampling}~\label{subsec:star}
Because our framework deforms real reference pixels rather than hallucinating them, its synthesis quality is inherently bounded by the quality of the reference pool: if the selected references lack the phonetic textures demanded by the target audio, the deformation has no faithful exemplar to follow and degenerates into stretched artifacts or blurred intra-oral regions. Conventional random sampling aggravates this, as it tends to return homogeneous topologies and occasional motion-blurred frames. We therefore propose Sharpness and Topology-Aware Reference Sampling (STAR Sampling), a deterministic procedure that selects a reference pool which is both topologically diverse and visually sharp. As it runs only once before inference, STAR Sampling adds no latency to synthesis.

\noindent\textbf{Feature Space Construction.} To select references by their articulatory state rather than by raw appearance, we represent each frame as a set of geometry-aware attribute. As shown in Fig.~\ref{figs:topo}, we first extract a few foundational measurements from each frame: the rigid facial width $W_{\text{face}}$, measured between cheek anchors; the inner-mouth height $h_{\text{in}}$ and width $w_{\text{in}}$; the inner oral-cavity area $A_{\text{mouth}}$; and the mean normalized intensity $I_{\text{mean}} \in [0,1]$ within the cavity. To remain invariant to subject scale across a sequence, all measurements are normalized by the facial width. The vertical aspect ratio and horizontal stretch describe the geometric state of the lips, $D_{\text{mar}} = h_{\text{in}} / W_{\text{face}}$ and $D_{\text{width}} = w_{\text{in}} / W_{\text{face}}$. To characterize internal phonetic detail, the teeth exposure and cavity depth are derived from the mean oral intensity and normalized by the squared facial width:
\begin{equation}
    D_{\text{teeth}} = I_{\text{mean}} \cdot \frac{A_{\text{mouth}}}{W_{\text{face}}^2}, \qquad
    D_{\text{cavity}} = (1 - I_{\text{mean}}) \cdot \frac{A_{\text{mouth}}}{W_{\text{face}}^2},
    \label{eq:star_texture}
\end{equation}
where a high $D_{\text{teeth}}$ indicates exposed bright teeth and a high $D_{\text{cavity}}$ a deep, dark cavity. The fifth dimension is an image sharpness term that penalizes motion blur, defined as the variance of the Laplacian of the cropped face image, $D_{\text{sharp}} = \mathrm{VL}(I_{\text{face}})$.

\begin{algorithm}[th!]
\caption{STAR Sampling Algorithm}
\label{alg:star_fps}
\textbf{Input:} Topology features $\mathbf{F}_\text{topo} \in \mathbb{R}^{M \times 4}$, Sharpness metrics $\mathbf{D}_{\text{sharp}} \in \mathbb{R}^M$, Target anchor count $N$, Neighborhood size $k$ \\
\textbf{Output:} Selected reference frame indices $\mathcal{M}$
\begin{algorithmic}[1]
\STATE Standardize $\mathbf{F}_\text{topo}$ using Z-score normalization.
\STATE \textit{\# 1. Local Sharpness Normalization}
\FOR{each frame $i \in \{1, \dots, M\}$}
    \STATE Find the $k$ nearest neighbors of $i$ in the $\mathbf{F}_\text{topo}$ space.
    \STATE Let $d_{min}$ and $d_{max}$ be the minimum and maximum $D_{\text{sharp}}$ in this local neighborhood.
    \IF{$d_{max} > d_{min}$}
        \STATE $\hat{D}_{\text{sharp}, i} \leftarrow 0.1 + 0.9 \times \frac{D_{\text{sharp}, i} - d_{min}}{d_{max} - d_{min}}$ 
    \ELSE
        \STATE $\hat{D}_{\text{sharp}, i} \leftarrow 1.0$
    \ENDIF
\ENDFOR

\STATE \textit{\# 2. Anchor Initialization}
\STATE Compute distance to mean topology: $D_{mean, i} \leftarrow \|\mathbf{F}_{\text{topo}, i} - \mu_\text{topo}\|_2$
\STATE Select the first anchor $a \leftarrow \arg\max_i (\hat{D}_{\text{sharp}, i} / (D_{mean, i} + \epsilon))$
\STATE Initialize selected set $\mathcal{M} \leftarrow \{a\}$
\STATE Initialize shortest distance array $\mathcal{D} \leftarrow \infty$ for all $M$ frames, $\mathcal{D}_a \leftarrow 0$

\STATE \textit{\# 3. Sharpness-Weighted Iterative Sampling}
\WHILE{$|\mathcal{M}| < N$}
    \FOR{each unselected frame $i \notin \mathcal{M}$}
        \STATE Update shortest distance to selected set: \\
               $\mathcal{D}_i \leftarrow \min(\mathcal{D}_i, \|\mathbf{F}_{\text{topo}, i} - \mathbf{F}_{\text{topo}, a}\|_2)$
        \STATE Compute final weighted score: $Score_i \leftarrow \mathcal{D}_i \times \hat{D}_{\text{sharp}, i}$
    \ENDFOR
    \STATE Select next frame $a \leftarrow \arg\max_i Score_i$
    \STATE $\mathcal{M} \leftarrow \mathcal{M} \cup \{a\}$
    \STATE $\mathcal{D}_a \leftarrow 0$
\ENDWHILE
\RETURN $\mathcal{M}$
\end{algorithmic}
\label{alg:star}
\end{algorithm}

% \noindent\textbf{Sharpness-Aware Farthest-Point Sampling.} To select anchors that maximize diversity, we decouple the feature space into a four-dimensional topological subspace $\mathbf{F}_{\text{topo}} \in \mathbb{R}^{4}$ and the one-dimensional sharpness score $D_{\text{sharp}}$, and perform farthest-point sampling (FPS) over $\mathbf{F}_{\text{topo}}$. Naively, however, FPS may favor motion-blurred frames whose distorted topologies appear deceptively unique, while a global sharpness comparison is itself misleading, since an open mouth with visible teeth naturally yields a higher Laplacian variance than a closed one. We therefore normalize each frame's sharpness only against its $k$ nearest neighbors in $\mathbf{F}_{\text{topo}}$, obtaining a relative clarity score $\hat{D}_{\text{sharp}} \in (0,1]$, and weight the topological FPS distance by this relative clarity, as detailed in Algorithm~\ref{alg:star}. This penalizes genuinely blurred frames while fairly retaining rare but informative lip postures. In practice, we first draw $M = 100$ candidate frames at random to balance efficiency and coverage, and set the neighborhood size to $k = M / N$.

\begin{figure*}[t]
    \centering
    \includegraphics[width=\textwidth]{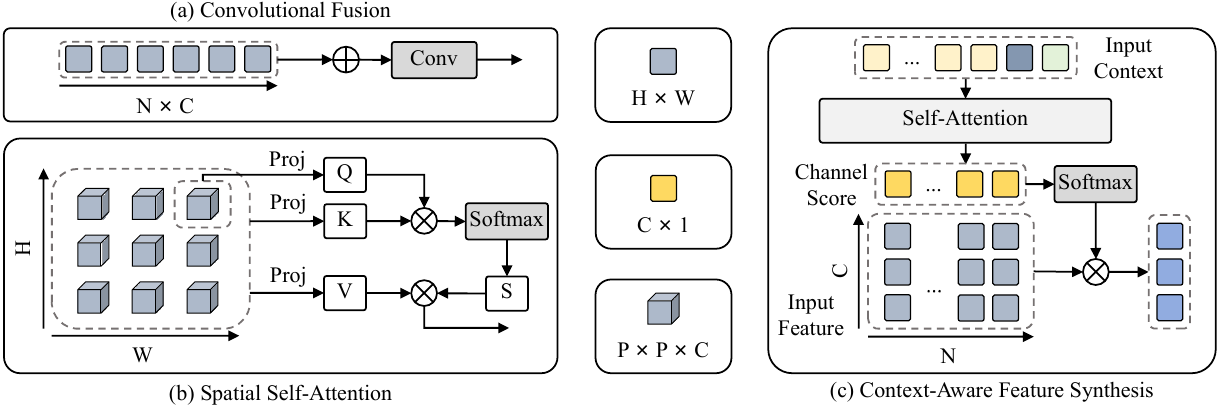}
    \caption{\textbf{Comparison of feature fusion mechanisms.} (a) Convolutional Fusion aggregates concatenated reference features using standard convolutional layers. (b) Spatial Self-Attention computes attention weights based on flattened spatial patches. (c) Our proposed \textbf{Context-Aware Feature Synthesis (CAFS)} leverages global context to predict channel-wise routing scores, aggregating diverse candidate features through an efficient channel-wise weighted summation without spatial or channel entanglement.}
    \label{figs:cafs}
\end{figure*}

\noindent\textbf{Sharpness-Aware Farthest-Point Sampling.} We first construct a four-dimensional topological subspace $\mathbf{F}_{\text{topo}} = [D_{\text{mar}}, D_{\text{width}}, D_{\text{teeth}}, D_{\text{cavity}}] \in \mathbb{R}^{4}$ and the one-dimensional sharpness score $D_{\text{sharp}}$, and select $N$ anchors through the sharpness-weighted farthest-point sampling  summarized in Algorithm~\ref{alg:star} and Fig,~\ref{figs:star}. The procedure consists of three steps, after $\mathbf{F}_{\text{topo}}$ is standardized by Z-score normalization so that all topological dimensions contribute comparably to the distance metric.

\textit{Step 1: Local sharpness normalization.} A global comparison of $D_{\text{sharp}}$ is misleading, since an open mouth with visible teeth naturally yields a higher Laplacian variance than a closed one. We therefore normalize each frame's $D_{\text{sharp}}$ only against its $k$ nearest neighbors in $\mathbf{F}_{\text{topo}}$, obtaining a relative clarity score $\hat{D}_{\text{sharp}} \in [0.1, 1]$. The lower bound of $0.1$ guarantees that even the blurriest frame retains a small chance of selection when it is topologically rare, preventing permanent exclusion.

\textit{Step 2: Anchor initialization.} The first anchor should be both clear and representative, so we select the frame that maximizes relative clarity while lying closest to the topological centroid $\mu_{\text{topo}}$:
\begin{equation}
    a \leftarrow \arg\max_{i}\ \frac{\hat{D}_{\text{sharp},i}}{D_{\text{mean},i} + \epsilon},
    \quad D_{\text{mean},i} = \lVert \mathbf{F}_{\text{topo},i} - \mu_{\text{topo}} \rVert_2,
    \label{eq:star_init}
\end{equation}
where $D_{\text{mean},i}$ is the distance of frame $i$ to the centroid and $\epsilon$ avoids division by zero. This anchor initializes the selected set $\mathcal{M}$, together with a shortest-distance array $\mathcal{D}$ used in the next step.

\textit{Step 3: Sharpness-weighted iterative sampling.} Each subsequent anchor should instead be clear yet maximally diverse from $\mathcal{M}$. We thus weight the FPS distance $\mathcal{D}_i = \min_{j \in \mathcal{M}} \lVert \mathbf{F}_{\text{topo},i} - \mathbf{F}_{\text{topo},j} \rVert_2$ by the same relative clarity and select
\begin{equation}
    a \leftarrow \arg\max_{i \notin \mathcal{M}}\ \mathcal{D}_i \cdot \hat{D}_{\text{sharp},i},
    \label{eq:star_iter}
\end{equation}
repeating until $|\mathcal{M}| = N$. The opposite roles of $D_{\text{mean}}$ (minimized at initialization) and $\mathcal{D}$ (maximized thereafter) let the first anchor capture a typical posture while the rest expand coverage toward rare but informative ones, with the clarity factor consistently penalizing blurred candidates in both phases. In practice, we draw $M = 100$ candidate frames at random to balance efficiency and coverage, and set the neighborhood size to $k = M / N$.

\begin{figure*}[ht]
    \centering
    \includegraphics[width=\textwidth]{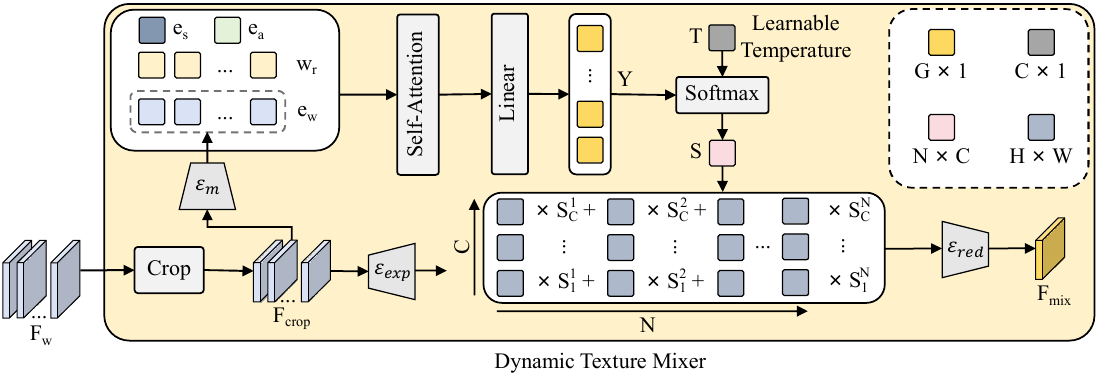}
    \caption{\textbf{Architecture of the Dynamic Texture Mixer.} The cropped reference features are encoded and fused with the global context through a Self-Attention block. Linear mapping and Softmax operations translate the updated tokens into channel-specific routing scores, under which the reference features undergo channel-wise weighted aggregation to synthesize the target mouth feature.}
    \label{figs:dtm}
\end{figure*}

\subsection{Texture Fusion via the Dynamic Texture Mixer}
The Spatial Alignment Module aligns the spatial structure of multiple references to the target pose, but this structural foundation alone is insufficient: owing to the dynamic deformation of mouth shapes and the transience of facial textures, no single warped reference can reconstruct rich, high-fidelity detail. It is therefore necessary to fuse complementary texture cues across the references without destroying the very textures that the deformation paradigm is meant to preserve.

As illustrated in Fig.~\ref{figs:cafs}, we categorize multi-reference feature fusion into three mechanisms. The first two are widely adopted in existing methods: convolutional fusion~\cite{zhang2023dinet,liu2026identity} (Fig.~\ref{figs:cafs}(a)) and spatial self-attention~\cite{zhang2024musetalk,li2024latentsync} (Fig.~\ref{figs:cafs}(b)). As analyzed below, both inadvertently corrupt the reference textures. We therefore propose a third mechanism, Context-Aware Feature Synthesis (CAFS) (Fig.~\ref{figs:cafs}(c)), which casts fusion as a channel-wise selection problem and thereby keeps the references intact. To adapt this mechanism to the lip-sync task, we instantiate it as the Dynamic Texture Mixer (DTM).

\noindent\textbf{Convolutional Fusion.} Convolutional fusion concatenates references along the channel dimension and processes them with multi-layer convolutions. It has two intrinsic drawbacks. First, fixed convolutional weights cannot adapt the fusion to the driving conditions; even with conditional injection (e.g., AdaIN) or channel attention~\cite{hu2018squeeze}, the underlying cross-reference and cross-channel convolutions still induce channel coupling, mixing independent features and averaging out high-frequency detail. Second, capturing global context requires stacking convolutions to enlarge the receptive field, and this repeated spatial aggregation further smooths textures while raising cost. Concretely, its complexity is $O(N \cdot K^2 \cdot H \cdot W \cdot C^2)$, where $N$, $K$, and $C$ are the number of references, the kernel size, and the channel dimension; since $K$ is small, this simplifies to $O(N \cdot H \cdot W \cdot C^2)$. The tight coupling of spatial size, reference count, and dense channel interaction yields a heavy overhead.

\noindent\textbf{Spatial Self-Attention.} Spatial self-attention, commonly employed within denoising networks, partitions the feature map into patches of size $P \times P \times C$, each linearly projected into Query, Key, and Value for standard attention; in essence, it relates the full channel vector at one location to those at all other locations and updates features by a weighted sum. This scheme has three limitations. First, it shares attention weights across channels, applying an identical per-location weight to all channels and thereby ignoring their distinct spatial semantics. Second, partitioning the feature map into patches fractures the continuous textures, degrading high-frequency detail. Third, attending over all spatial patches forms a large attention matrix; for instance, with patch size $P=1$ as in LatentSync~\cite{li2024latentsync}, the complexity grows quadratically with spatial resolution, reaching $O((HW)^2 C)$ and incurring substantial memory and compute costs.

\noindent\textbf{Context-Aware Feature Synthesis.} In summary, convolutional fusion erases high-frequency features through its fixed weights and linear summation, while spatial self-attention destroys reference texture priors and is too costly for real-time synthesis. Our third mechanism stems from a key observation about the features after spatial alignment. Because all references are processed by shared convolutional layers, identical channels carry the same visual semantics; moreover, after the Spatial Alignment Module their spatial structures are highly aligned. Consequently, the references collectively provide a rich pool of candidate features at the same spatial locations within identical semantic channels, and fusion reduces to a selection task: for each semantic channel, choosing the most suitable candidate given the current context. Following this insight, CAFS uses a self-attention module to capture the global context, maps each candidate's context-aware token into channel-specific routing weights through linear layers, and aggregates the candidates by channel-wise weighted summation. Crucially, it never injects conditions to modulate the image features themselves, so the original channel semantics, texture integrity, and spatial structure of the references are preserved intact. This selection-based design is also markedly more efficient: the self-attention and linear scoring take $\mathcal{O}(N^2 \cdot d + N \cdot C \cdot d)$ operations, where $d$ is the token dimension, and the channel-wise aggregation takes $\mathcal{O}(N \cdot H \cdot W \cdot C)$. Since the spatial resolution $H \times W$ typically dominates both $N$ and $d$, the overall complexity simplifies to $\mathcal{O}(N \cdot H \cdot W \cdot C)$, markedly lower than convolutional fusion ($\mathcal{O}(N \cdot H \cdot W \cdot C^2)$) and spatial self-attention ($\mathcal{O}((HW)^2 \cdot C)$).

\noindent\textbf{Dynamic Texture Mixer.} We instantiate CAFS as the Dynamic Texture Mixer (DTM) to mix the set of warped reference features $\mathbf{F}_{w} = \{\mathbf{F}_{w}^{n}\}_{n=1}^{N}$ produced by SAM (as shown in Fig.~\ref{figs:dtm}). Since SAM operates on full-face references ($H \times W$, with $H > W$) to align them with the source, their upper-face region becomes redundant for mouth synthesis and would only waste computation. We therefore crop each $\mathbf{F}_{w}^{n}$ along the height dimension down to $W$, retaining a square $W \times W$ lower-face region denoted $\mathbf{F}_{crop}^{n}$.

The DTM then proceeds along two parallel branches. The first prepares the candidates to be mixed: an expansion network $\mathcal{E}_{exp}$ lifts each $\mathbf{F}_{crop}^{n}$ into a higher channel dimension $C'$ to enlarge the representational capacity for aggregation,
\begin{equation}
    \mathbf{F}_{exp}^{n} = \mathcal{E}_{exp}(\mathbf{F}_{crop}^{n}) \in \mathbb{R}^{W \times W \times C'}, \quad \forall n \in [1,N].
    \label{eq:dtm_expand}
\end{equation}
The second branch prepares the routing condition: a condition encoder $\mathcal{E}_m$ summarizes each $\mathbf{F}_{crop}^{n}$ into a single token $\mathbf{e}_{w}^{n}$ that conveys the current state of that reference. We concatenate these tokens with the context carried over from SAM, namely $\{\mathbf{w}_{r}^{i}\}_{i=1}^{N}$, $\mathbf{e}_{s}$, and $\mathbf{e}_{a}$, and pass them through a multi-head self-attention block; the output tokens $\{\mathbf{m}_{w}^{n}\}_{n=1}^{N}$ corresponding to $\{\mathbf{e}_{w}^{n}\}$ thus encode each reference's relevance under the global driving context. A linear layer $\mathcal{F}_{proj}$ then maps each $\mathbf{m}_{w}^{n}$ into a group-wise routing score:
\begin{equation}
    \mathbf{Y}^{n} = \mathcal{F}_{proj}(\mathbf{m}_{w}^{n}) \in \mathbb{R}^{G}, \quad \forall n \in [1,N],
    \label{eq:dtm_score}
\end{equation}
where $G \ll C'$ is the number of channel groups. Predicting one score per group, rather than per channel, avoids regressing $N \times C'$ free weights and reflects the prior that strongly correlated channels should share a routing decision.

To recover channel-wise weights, the group score $\mathbf{Y}_{g}^{n}$ is replicated across all channels of its group $g$ to form $\mathbf{Y}_{c}^{n}$. While this keeps the coarse decision consistent within a group, identical weights would over-constrain the aggregation; we therefore attach a learnable per-channel temperature $\mathbf{T}_{c}$ that lets each channel sharpen or soften the shared score independently. The final aggregation weight of the $n$-th reference at channel $c$ is obtained by a Softmax over the $N$ references:
\begin{equation}
    \mathbf{S}_{c}^{n} = \frac{\exp(\mathbf{Y}_{c}^{n} / \mathbf{T}_{c})}{\sum_{i=1}^{N} \exp(\mathbf{Y}_{c}^{i} / \mathbf{T}_{c})}, \quad \forall c \in [1,C'],\ \forall n \in [1,N].
    \label{eq:dtm_softmax}
\end{equation}
Stacking $\mathbf{S}_{c}^{n}$ over channels gives the per-reference weight vector $\mathbf{S}^{n} \in \mathbb{R}^{C'}$. Finally, we aggregate the expanded candidates by a channel-wise weighted summation and project the result back to the original channel size through a reduction network $\mathcal{E}_{red}$, yielding the mixed mouth feature $\mathbf{F}_{mix}$:
\begin{equation}
    \mathbf{F}_{mix} = \mathcal{E}_{red}\!\left(\sum_{n=1}^{N} \mathbf{S}^{n} \odot \mathbf{F}_{exp}^{n}\right),
    \label{eq:dtm_mix}
\end{equation}
where $\odot$ denotes channel-wise multiplication broadcast across the spatial dimensions. Because each output channel is a convex combination of the corresponding channel across references, the DTM selects rather than synthesizes, leaving the spatial structure and high-frequency texture of every reference intact.

\subsection{Spatio-Temporal Shifted Adaptive Masking}
As discussed in Sec.~\ref{sec:intro}, prevailing mask strategies face a dilemma between leakage suppression and background fidelity, whose root cause is treating the source background as a fixed, immutable environment. We instead decouple the source frame into global attributes that condition lip generation and a background prior that is independent of the lip shape, and realize this idea through the Spatio-Temporal Shifted Adaptive Masking (STAM) strategy. Lip generation is conditioned solely on the source frame under a fixed large mask, which entirely deprives the network of real visual hints about the lower face, while the background prior is supplied by a separate background reference $\mathbf{I}_{BG}$.

The key to STAM lies in how this background reference is constructed during training versus inference. During training, rather than using the source frame itself, we take an adjacent frame at a small temporal offset as the background reference. The rationale is that, within a short window, the head moves little, so the surrounding background and torso change only marginally and still constitute a high-fidelity texture reference; yet the lip shape of this shifted frame differs from the target, so its chin position is misaligned. Confronted with this mismatch, the model cannot trivially copy pixels and is instead forced to learn realistic elastic stretching and redrawing of the background skin and jawline in accordance with the newly generated lip. During inference, the background reference reduces to the current source frame, that is, a zero temporal offset, which preserves the genuine background exactly while the model, trained under the stricter shifted regime, retains the spatial adaptivity needed to stretch the chin and blend the audio-driven mouth seamlessly into the surroundings, eliminating double chins and boundary seams.

Concretely, a background encoder $\mathcal{E}_{bg}$ extracts a background feature $\mathbf{F}_{bg}$ from $\mathbf{I}_{BG}$. Since $\mathbf{F}_{mix}$ covers only the cropped mouth region, we zero-pad it back to the full-face spatial size, denoted $\hat{\mathbf{F}}_{mix}$, and concatenate it with the source feature $\mathbf{F}_{s}$ and the background feature $\mathbf{F}_{bg}$ to predict the background motion flow:
\begin{equation}
    \mathbf{M}_{bg} = \mathcal{F}_{conv}(\hat{\mathbf{F}}_{mix} \oplus \mathbf{F}_{s} \oplus \mathbf{F}_{bg}).
    \label{eq:bg_flow}
\end{equation}
The background feature is then warped by $\mathbf{M}_{bg}$ and cropped to the mouth-region size, yielding the aligned background feature $\hat{\mathbf{F}}_{bg} = \text{Crop}(\text{Warp}(\mathbf{F}_{bg}, \mathbf{M}_{bg}))$. Finally, $\mathbf{F}_{mix}$ and $\hat{\mathbf{F}}_{bg}$ are concatenated along the channel dimension and decoded by a SPADE decoder into the synchronized output face:
\begin{equation}
    \mathbf{I}_{o} = \mathcal{F}_{spade}(\mathbf{F}_{mix} \oplus \hat{\mathbf{F}}_{bg}).
    \label{eq:spade_decode}
\end{equation}

\subsection{Loss Function}
We train the model with a composite objective. Let $\mathbf{I}_{o}$ and $\mathbf{I}_{gt}$ denote the output image and the corresponding ground-truth frame. To encourage realistic images with sharp high-frequency detail, we adopt the hinge adversarial loss $\mathcal{L}_{gan}$~\cite{lim2017geometric}, comprising the standard discriminator and generator terms. A pre-trained VGG network~\cite{simonyan2014very} provides a multi-scale perceptual loss $\mathcal{L}_{perc}$ that enforces global structural consistency, computed as the $\ell_1$ distance between the activation maps of $\mathbf{I}_{o}$ and $\mathbf{I}_{gt}$. To preserve fine-grained intra-oral detail, we apply an LPIPS loss~\cite{zhang2018unreasonable} $\mathcal{L}_{lpips}$ restricted to the central mouth region of the generated image. To enforce strict audio-visual synchronization, a pre-trained SyncNet~\cite{prajwal2020lip} yields a sync loss $\mathcal{L}_{sync}$ over each window of 16 consecutive frames and the corresponding audio. We further incorporate the TREPA loss~\cite{li2024latentsync} $\mathcal{L}_{trepa}$, measured as the squared distance between the embeddings of generated and ground-truth clips under a self-supervised video encoder, to improve the temporal naturalness of the synthesized textures. The total objective is the weighted sum of these terms:
\begin{equation}
    \mathcal{L}_{total} = \lambda_1 \mathcal{L}_{gan} + \lambda_2 \mathcal{L}_{perc} + \lambda_3 \mathcal{L}_{lpips} + \lambda_4 \mathcal{L}_{sync} + \lambda_5 \mathcal{L}_{trepa},
    \label{eq:loss_total}
\end{equation}
where $\lambda_1, \dots, \lambda_5$ are the corresponding weighting hyperparameters.

\section{Experiments}\label{sec:experiments}

\begin{figure*}[t]
    \centering
    \includegraphics[width=\textwidth]{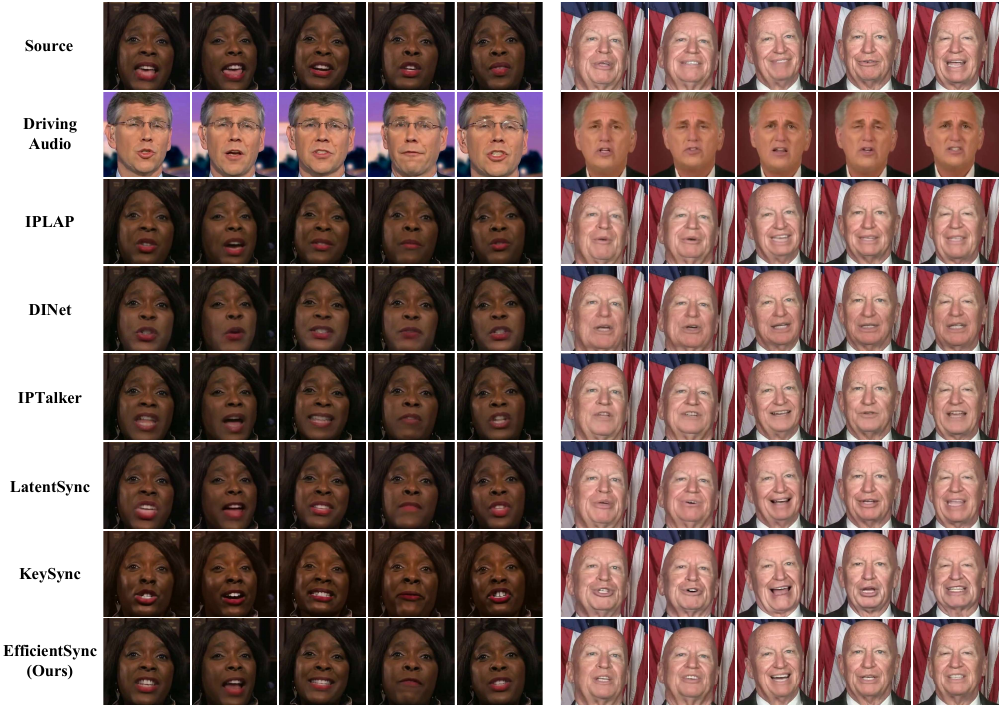}
    \caption{\textbf{Qualitative comparisons with state-of-the-art works under the cross-audio setting.} EfficientSync preserves sharp, identity-consistent intra-oral textures, whereas competing methods either blur the mouth region or hallucinate teeth that deviate from the source identity.}
    \label{figs:vssota}
\end{figure*}

\noindent\textbf{Datasets.} We conduct experiments on two high-resolution talking-face datasets, HDTF~\cite{zhang2021flow} and VFHQ~\cite{xie2022vfhq}. HDTF comprises 362 high-definition videos at resolutions ranging from roughly 720p to 1080p, while VFHQ contains about 8{,}000 videos spanning more than 20 countries and a wide variety of languages. For evaluation, we randomly select 20 videos from the test set of each dataset.

\noindent\textbf{Implementation Details.} All videos are resampled to 25 fps, and 468 facial landmarks are extracted with MediaPipe~\cite{lugaresi2019mediapipe}. Based on these landmarks, the facial region is cropped and resized to a $416 \times 320$ resolution, and the audio is resampled to 16 kHz. During training, the number of reference frames $N$ is randomly sampled from $[3, 7]$.

\noindent\textbf{Evaluation Metrics.} We evaluate our method along three dimensions. \emph{(1) Visual quality.} We measure structural fidelity with SSIM~\cite{wang2004image} and perceptual similarity with LPIPS~\cite{zhang2018unreasonable}, assess overall frame-level and video-level quality with the Fr\'echet Inception Distance (FID)~\cite{heusel2017gans} and Fr\'echet Video Distance (FVD)~\cite{unterthiner2018towards}, respectively, and quantify image blurriness with VL~\cite{pech2000diatom}. \emph{(2) Lip-sync accuracy.} We adopt $\text{Sync}_{\text{conf}}$~\cite{chung2016out} to gauge audio-visual synchronization. \emph{(3) Efficiency.} To reflect practical deployability, we report inference speed (FPS), peak memory usage (Mem, in MB), and floating-point operations per frame (GFLOPs/f).

\subsection{Quantitative Evaluation}
We compare EfficientSync against recent state-of-the-art methods, namely IPLAP~\cite{zhong2023identity}, DINet~\cite{zhang2023dinet}, LatentSync~\cite{li2024latentsync}, IPTalker~\cite{liu2026identity}, and KeySync~\cite{bigata2025keysync}, under both the reconstruction and cross-audio settings. The results are reported in Tables~\ref{tab:quantitative_results_reconstruction} and~\ref{tab:quantitative_results_cross}, where the best and second-best scores are highlighted in bold and underlined, respectively. For all comparisons, the reference frame count of our model is fixed to 5, which is a common setting in previous multi-reference models.

\noindent\textbf{Visual quality.} As shown in Table~\ref{tab:quantitative_results_reconstruction}, EfficientSync attains the best reconstruction quality on the majority of visual metrics across both datasets. On HDTF it reaches the highest SSIM (0.979), LPIPS (0.013), VL (58.69), FID (1.82), and FVD (49.14), improving the strongest baseline LatentSync on FID by 22\% (1.82 vs. 2.34) and on VL by 22\% (58.69 vs. 47.91). The same trend holds on VFHQ, where it leads on SSIM (0.967), LPIPS (0.022), and FID (4.13), more than halving the FID of LatentSync (4.13 vs. 5.78). These consistent gains in sharpness-sensitive metrics (VL, LPIPS) confirm that deforming real reference pixels preserves fine intra-oral texture far better than generative resynthesis. The cross-audio results in Table~\ref{tab:quantitative_results_cross} echo this conclusion: EfficientSync obtains the best VL and FID on both datasets, for instance reducing the HDTF FID to 1.92 against LatentSync's 2.47.

\noindent\textbf{Lip-sync accuracy.} On $\text{Sync}_{\text{conf}}$, our method occupies a competitive mid-tier position rather than the top. Under cross-audio on HDTF it scores 7.59, trailing the diffusion-based LatentSync (9.04) and KeySync (8.55) but clearly surpassing the warping-based IPTalker (4.62), DINet (7.36), and IPLAP (5.86). This gap is expected: diffusion backbones optimize synchronization through a powerful generative prior at the cost of identity and speed, whereas our deformation paradigm prioritizes texture fidelity. Notably, EfficientSync still achieves the strongest synchronization among all deformation- and GAN-based methods, indicating that the accuracy trade-off is modest and well compensated by its substantial gains in fidelity and efficiency.

\begin{table}[t]
\centering
\caption{Comparison of computational efficiency on a single NVIDIA A100 GPU.}
\label{tab:efficiency_results}
\setlength{\tabcolsep}{2.0mm}
\begin{tabular}{l ccc}
\toprule
Method & FPS $\uparrow$ & Mem (MB) $\downarrow$ & GFLOPs/f $\downarrow$ \\
\midrule
IPLAP\cite{zhong2023identity} & 0.98 & 8848 & 2356 \\
DINet\cite{zhang2023dinet} & 10.73 & \textbf{349} & \textbf{180}  \\
IPTalker\cite{liu2026identity} & \underline{27.21} & \underline{565} & 366 \\
LatentSync\cite{li2024latentsync} & 4.39 & 11704 & 39840  \\
KeySync\cite{bigata2025keysync} & 1.07 & 38363 & 16358  \\
\midrule
EfficientSync (Ours) & \textbf{166.01} & 718 & \underline{245}  \\
\bottomrule
\end{tabular}
\end{table}

\noindent\textbf{Efficiency.} Table~\ref{tab:efficiency_results} reports efficiency on a single NVIDIA A100 GPU. EfficientSync runs at 166.01 FPS, over $6\times$ faster than the quickest prior method IPTalker (27.21 FPS) and roughly $38\times$ faster than the diffusion-based LatentSync (4.39 FPS), while consuming only 245 GFLOPs per frame, two orders of magnitude below LatentSync (39{,}840) and KeySync (16{,}358), and 718 MB of peak memory against their 11.7 GB and 38.4 GB. This efficiency is not incidental but a direct consequence of our design. First, unlike diffusion baselines that reconstruct the entire face from noise through tens of denoising steps, EfficientSync edits only the lower-face region in a single forward pass, avoiding both the iterative sampling that dominates their latency and the redundant regeneration of the unchanged upper face. Second, the Dynamic Texture Mixer replaces spatial self-attention, whose cost grows quadratically with resolution as $O((HW)^2 C)$, with a channel-wise selection whose cost is linear in the spatial size, $O(N \cdot H \cdot W \cdot C)$; this is the primary source of our low GFLOPs and memory, as it never materializes the large spatial attention matrices that the diffusion methods rely on. Third, the reference flows are predicted and applied in parallel, so enlarging the reference pool adds little latency. Built entirely from lightweight convolutional and linear operators rather than a heavy VAE-UNet denoiser, EfficientSync remains deployable under constrained hardware, a regime in which the diffusion competitors are impractical.

% reconstruction / cross-audio tables unchanged
\begin{table*}[ht]
\centering
\caption{Quantitative comparison of reconstruction performance on the HDTF and VFHQ datasets.}
\label{tab:quantitative_results_reconstruction}
\setlength{\tabcolsep}{1mm}
\begin{tabular}{l cccccc cccccc}
\toprule
\multirow{2}{*}{Method} & \multicolumn{6}{c}{HDTF} & \multicolumn{6}{c}{VFHQ} \\
\cmidrule(lr){2-7} \cmidrule(lr){8-13}
& SSIM $\uparrow$ & LPIPS $\downarrow$ & VL $\uparrow$ & FID $\downarrow$  & FVD $\downarrow$ & $\text{Sync}_{\text{conf}}$ $\uparrow$ & SSIM $\uparrow$ & LPIPS $\downarrow$ & VL $\uparrow$ & FID $\downarrow$ & FVD $\downarrow$ & $\text{Sync}_{\text{conf}}$ $\uparrow$ \\
\midrule
IPLAP\cite{zhong2023identity} & 0.958 & 0.032 & 37.615 & 3.895 & 95.104 & 8.039 & 0.942 & 0.057 & 61.359 & 9.451 & 117.370 & 5.751 \\
DINet\cite{zhang2023dinet} & 0.952 & 0.030 & 47.354 & 3.690 & 94.625 & 8.851 & 0.925 & 0.071 & 59.505 & 20.399 & 260.732 & 5.140 \\
IPTalker\cite{liu2026identity} & \underline{0.971} & \underline{0.018} & \underline{51.750} & 4.433 & 61.798 & 8.885 & 0.945 & \underline{0.040} & \textbf{85.791} & 9.365 & 167.655 & 5.530 \\
LatentSync\cite{li2024latentsync} & 0.963 & 0.028 & 47.906 & \underline{2.342} & \underline{50.820} & \textbf{9.183} & \underline{0.952} & 0.042 & 68.544 & \underline{5.782} & \textbf{54.241} & \textbf{7.765} \\
KeySync\cite{bigata2025keysync} & 0.906 & 0.064 & 42.868 & 11.321 & 121.321 & 8.765 & 0.878 & 0.097 & 74.002 & 23.421 & 149.354 & \underline{7.747} \\
\midrule
EfficientSync (Ours) & \textbf{0.979} & \textbf{0.013} & \textbf{58.691} & \textbf{1.824} & \textbf{49.144} & \underline{9.107} & \textbf{0.967} & \textbf{0.022} & \underline{85.410} & \textbf{4.133} & \underline{88.839} & 6.058 \\
\bottomrule
\end{tabular}
\end{table*}

\begin{table*}[t]
\centering
\caption{Quantitative comparison of cross-audio performance on the HDTF and VFHQ datasets.}
\label{tab:quantitative_results_cross}
\setlength{\tabcolsep}{2mm}
\begin{tabular}{l cccc cccc}
\toprule
\multirow{2}{*}{Method} & \multicolumn{4}{c}{HDTF} & \multicolumn{4}{c}{VFHQ} \\
\cmidrule(lr){2-5} \cmidrule(lr){6-9}
& VL $\uparrow$ & FID $\downarrow$  & FVD $\downarrow$ & $\text{Sync}_{\text{conf}}$ $\uparrow$ & VL $\uparrow$ & FID $\downarrow$ & FVD $\downarrow$ & $\text{Sync}_{\text{conf}}$ $\uparrow$ \\
\midrule
IPLAP\cite{zhong2023identity} & 37.595 & 4.052 & 110.420 & 5.862 & 61.534 & 18.925 & 159.798 & 4.476\\
DINet\cite{zhang2023dinet} & 46.039 & 3.660 & 118.889 & 7.355 & 59.229 & 11.112 & 286.577 & 4.369\\
IPTalker\cite{liu2026identity} & \underline{51.733} & 4.606 & 87.501 & 4.622 & \underline{90.774} & 10.260 & 201.464 & 3.231\\
LatentSync\cite{li2024latentsync}& 47.714 & \underline{2.470} & \textbf{78.510} & \textbf{9.037} & 70.346 & \underline{7.745} & \textbf{149.811} & \underline{6.550}\\
KeySync\cite{bigata2025keysync} & 42.964 & 11.258 & 165.393 & \underline{8.547} & 79.739 & 25.876 & 258.167 & \textbf{6.989}\\
\midrule
EfficientSync (Ours)  & \textbf{57.331} & \textbf{1.916} & \underline{86.441} & 7.589 & \textbf{93.055} & \textbf{5.608} & \underline{151.852} & 4.494\\
\bottomrule
\end{tabular}
\end{table*}

\subsection{Qualitative Evaluation}
The cross-audio comparisons in Fig.~\ref{figs:vssota} corroborate the quantitative results. IPLAP and DINet produce visibly blurred mouth interiors, and although IPTalker and LatentSync sharpen the lips somewhat, a clear gap to the source textures remains. KeySync yields the crispest teeth but at the cost of fidelity, frequently deviating from the ground truth and producing exaggerated mouth shapes. EfficientSync instead strikes the best balance, rendering sharp, realistic lip motion whose intra-oral textures stay faithful to the reference identity and free of boundary artifacts.

\subsection{Ablation Study}
We ablate the two model-side designs, the Dynamic Texture Mixer (DTM) and the STAM strategy, on HDTF in Table~\ref{tab:ablation_study}.

\noindent\textbf{Feature fusion.} Replacing DTM with multi-layer convolutions (\textit{w/ Conv Fusion}) or spatial self-attention (\textit{w/ Attn Fusion}) degrades nearly all visual metrics: under reconstruction, FVD rises from 49.14 to 55.07 and 54.23 respectively, and FID from 1.82 to 2.06 and 2.28. The effect is far more pronounced under cross-audio, where the entanglement of differing reference mouth shapes in these operators corrupts lip motion: $\text{Sync}_{\text{conf}}$ drops from our 7.59 to 5.87 (Conv) and 6.91 (Attn). This confirms that the channel-wise selection in DTM, by keeping references disentangled, is essential not only for texture fidelity but also for accurate synchronization.

\begin{figure}[t]
    \centering
    \includegraphics[width=0.45\textwidth]{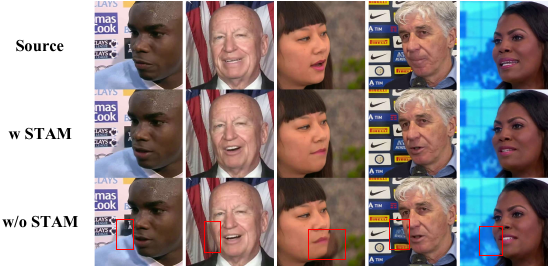}
    \caption{\textbf{Visual comparison for the STAM ablation.} As highlighted by the red boxes, removing STAM (w/o STAM) causes spatial discontinuities and blurred background around the lower face, whereas our full model (w STAM) preserves the background and blends seamlessly.}
    \label{figs:stam}
\end{figure}

\noindent\textbf{Mask strategy.} Removing the background reference and synthesizing directly from the fixed-mask source (\textit{w/o STAM}) degrades every visual metric, raising reconstruction FVD from 49.14 to 53.54 and LPIPS from 0.0127 to 0.0149. As Fig.~\ref{figs:stam} shows, this manifests as blurred background around the lower face and visible seams between the synthesized mouth and the surrounding region, whereas the full model preserves the background and blends seamlessly.

\begin{table}[t]
\centering
\caption{Ablation study on HDTF.}
\label{tab:ablation_study}
\setlength{\tabcolsep}{1mm}
\begin{tabular}{l cccccc}
\toprule
Variant & SSIM $\uparrow$ & LPIPS $\downarrow$ & VL $\uparrow$ & FID $\downarrow$ & FVD $\downarrow$ & $\text{Sync}_{\text{conf}}$ $\uparrow$ \\
\midrule
\multicolumn{7}{c}{\textit{Reconstruction}} \\
\midrule
w/ Conv Fusion & 0.9779 & 0.0132 & 57.281 & 2.055 & 55.071 & 8.947 \\
w/ Attn Fusion & 0.9786 & 0.0132 & 58.680 & 2.284 & 54.225 & 9.043 \\
w/o STAM & 0.9761 & 0.0149 & 56.900 & 1.945 & 53.541 & 8.670 \\
Ours & \textbf{0.9791} & \textbf{0.0127} & \textbf{58.691} & \textbf{1.824} & \textbf{49.144} & \textbf{9.107} \\
\midrule
\multicolumn{7}{c}{\textit{Cross-Audio}} \\
\midrule
w/ Conv Fusion & - & - & 57.275 & 2.319 & 91.750 & 5.869 \\
w/ Attn Fusion & - & - & \textbf{58.649} & 2.415 & 86.644 & 6.913 \\
w/o STAM & - & - & 56.840 & 2.347 & 88.847 & 7.346 \\
Ours & - & - & 57.331 & \textbf{1.916} & \textbf{86.441} & \textbf{7.589} \\
\bottomrule
\end{tabular}
\end{table}

\noindent\textbf{Effect of STAR Sampling.}
We also study STAR Sampling, which selects the reference pool described in Sec.~\ref{subsec:star}. Table~\ref{tab:refnum_star_ablation} compares it against random sampling and Farthest Point Sampling (FPS) under both reconstruction and cross-audio settings on HDTF, sweeping the number of references. Removing the sharpness-aware component from STAR Sampling effectively reduces it to FPS. Three observations stand out. First, introducing topological diversity via FPS significantly improves sampling efficiency and robustness. In reconstruction, FPS achieves a highly competitive $\text{Sync}_{\text{conf}}$ of 9.114 with only 8 references, whereas random sampling requires 16 to reach its peak. This advantage is more pronounced in cross-audio generation, where pure random selection struggles, while FPS preserves crucial extreme poses to maintain consistently higher scores. Second, incorporating sharpness awareness further elevates performance. In reconstruction, STAR Sampling reaches the highest overall $\text{Sync}_{\text{conf}}$ peak of 9.177, and in the cross-audio setting, it similarly attains the absolute peak of 7.569, showing that sharp structural priors consistently benefit lip-sync synthesis. Third, in the cross-audio setting, the peak performance for all three sampling methods consistently appears at exactly 4 references. Unlike reconstruction, which benefits from a larger reference pool (e.g., 8 or 16 frames) to capture fine-grained personal habits, cross-identity driving favors a more concise set of anchors. Supplying too many references in cross-driving likely introduces redundant, identity-specific topological nuances that interfere with the generalized audio-driven deformations, indicating that a compact but diverse reference pool is optimal for generalizable synthesis.

\begin{table}[t]
\centering
\caption{Ablation on the number of reference frames and STAR Sampling on HDTF.}
\label{tab:refnum_star_ablation}
\setlength{\tabcolsep}{1.0mm}
\begin{tabular}{l cccccc}
\toprule
\multirow{2}{*}{Ref. Num.} & \multicolumn{2}{c}{Random Sampling} & \multicolumn{2}{c}{FPS} & \multicolumn{2}{c}{STAR Sampling}\\
\cmidrule(lr){2-3} \cmidrule(lr){4-5} \cmidrule(lr){6-7}
& $\text{Sync}_{\text{conf}}$ $\uparrow$ & VL $\uparrow$ & $\text{Sync}_{\text{conf}}$ $\uparrow$ & VL $\uparrow$ & $\text{Sync}_{\text{conf}}$ $\uparrow$ & VL $\uparrow$\\
\midrule
\multicolumn{7}{c}{\textit{Reconstruction}} \\
\midrule
1    & 8.391 & 58.240 & 8.505 & 58.206 & 8.552 & 58.398 \\
2    & 8.771 & 58.538 & 8.833 & 58.579 & 8.910 & 58.703 \\
4    & 9.004 & 58.585 & 9.065 & \textbf{58.702} & 9.113 & \textbf{58.720} \\
8    & 9.076 & 58.651 & \textbf{9.114} & 58.669 & \textbf{9.177} & 58.680 \\
16   & \textbf{9.147} & 58.670 & 9.091 & 58.675 & 9.121 & 58.700 \\
32   & 9.078 & \textbf{58.672} & 9.079 & 58.666 & 9.077 & 58.680 \\
\midrule
Avg. & 8.911 & 58.559 & 8.948 & 58.583 & 8.992 & 58.647  \\
\midrule
\multicolumn{7}{c}{\textit{Cross-Audio}} \\
\midrule
1    & 6.299 & 56.732 & 6.543 & 56.814 & 6.592 & 56.909 \\
2    & 7.007 & 57.102 & 7.136 & 57.185 & 7.126 & 57.200 \\
4    & \textbf{7.483} & \textbf{57.277} & \textbf{7.558} & \textbf{57.276} & \textbf{7.569} & \textbf{57.272} \\
8    & 7.419 & 57.251 & 7.431 & 57.239 & 7.488 & 57.248 \\
16   & 6.978 & 57.268 & 7.301 & 57.220 & 7.244 & 57.262 \\
32   & 6.770 & 57.240 & 7.099 & 57.256 & 7.067 & 57.244 \\
\midrule
Avg. & 6.993 & 57.145 & 7.178 & 57.165 & 7.181 & 57.189 \\
\bottomrule
\end{tabular}
\end{table}

\subsection{User Study}

\begin{figure*}[ht]
    \centering
    \includegraphics[width=\textwidth]{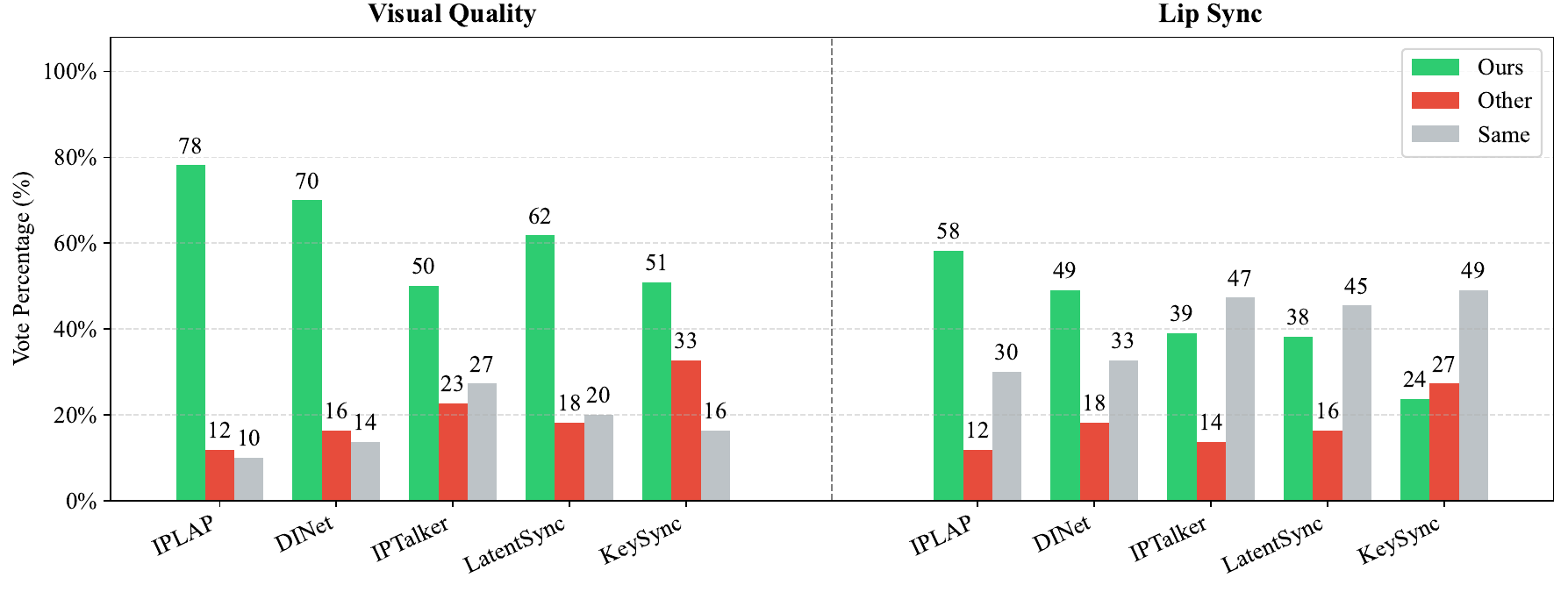}
    \caption{\textbf{User Preference Study.} We evaluated EfficientSync against random baselines through a user study focusing on Visual Quality and Lip Sync. The results display the preference rates as a percentage of total votes, indicating whether participants preferred our model (Ours), the competing model (Other), or viewed them as comparable (Same).}
    \label{figs:user_study}
\end{figure*}

We finally conduct a user study to compare EfficientSync against several baseline lip-sync models on videos generated from the test set. A total of 25 participants were presented with video pairs for each trial: one generated by our method and the other by a randomly selected baseline. Participants were asked to evaluate the videos based on visual quality and lip-sync accuracy, choosing the superior video or indicating that both were of equal quality. As shown in Fig.~\ref{figs:user_study}, users demonstrated a clear preference for EfficientSync in both visual quality and lip-sync accuracy over most baselines. The only exception was lip-sync accuracy against KeySync, where EfficientSync received slightly fewer votes; however, nearly half of the responses considered the two models to be comparable. Overall, the study highlights that EfficientSync achieves the best visual quality while maintaining highly competitive lip-sync accuracy.

\section{Analysis}\label{sec:analysis}
Beyond benchmarking against prior methods, we now examine \emph{why} EfficientSync works, focusing on how the Dynamic Texture Mixer (DTM) routes texture across references and how STAR Sampling shapes that routing. Unless otherwise noted, all analyses are conducted under the cross-audio setting on HDTF.
\begin{figure}[h]
    \centering
    \includegraphics[width=0.45\textwidth]{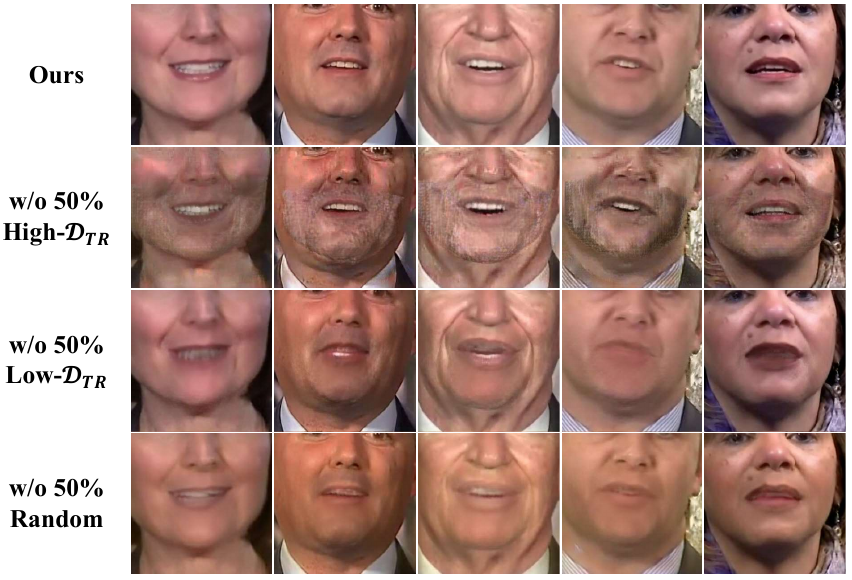}
    \caption{\textbf{Causal knockout experiments guided by Temporal Routing Divergence ($\mathcal{D}_{TR}$).} $\mathcal{D}_{TR}$-guided channel ablations reveal how DTM adaptively disentangles rigid high-frequency details from flexible low-frequency transitions.}
    \label{figs:TRD}
\end{figure}

\begin{figure*}[t]
    \centering
    \includegraphics[width=\textwidth]{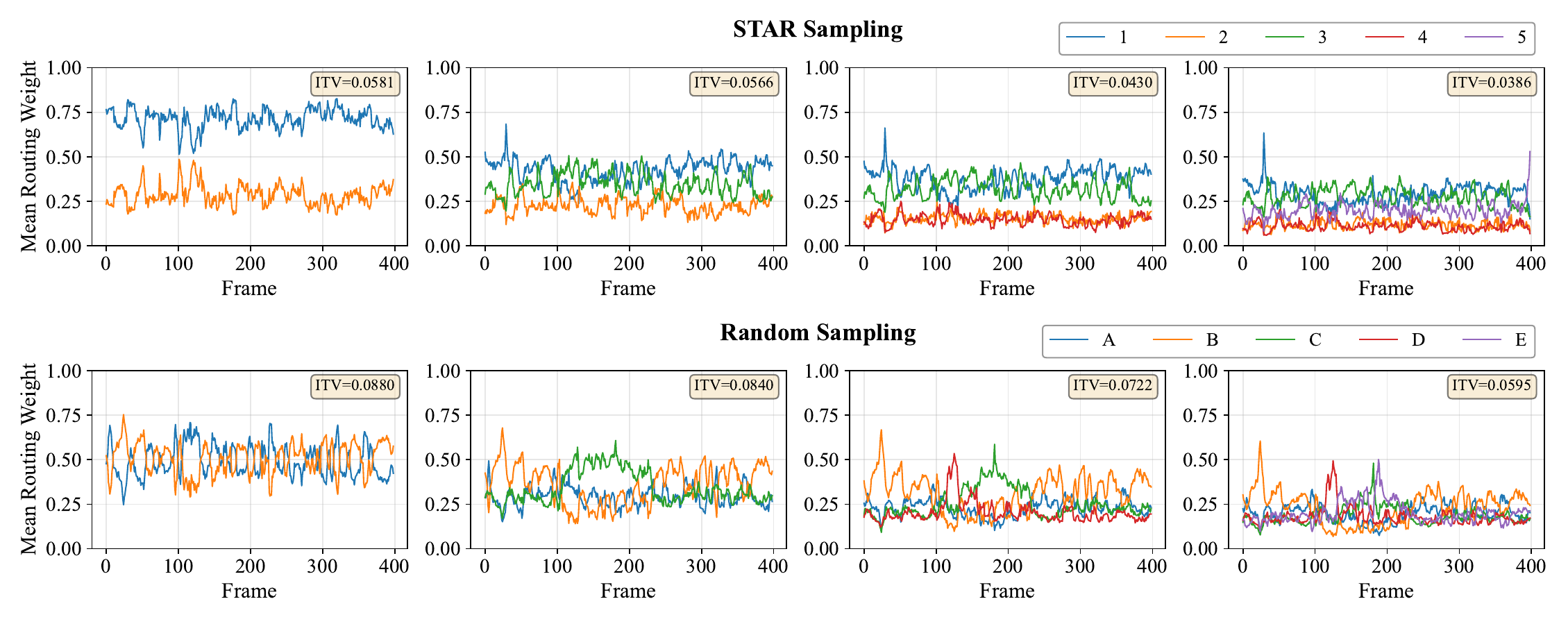}
    \caption{\textbf{Temporal trajectories of routing weights on a single video.} Visualization of the channel-averaged attention weights allocated by DTM to individual references under STAR and random sampling across varying reference counts.}
    \label{figs:comparison_routing}
\end{figure*}
\subsection{Channel-wise Routing Characteristics}
Recall that the DTM treats multi-reference fusion as a channel-wise routing process (Sec.~\ref{subsec:dtm}), akin to a mixing console that preserves texture integrity rather than homogenizing it as conventional convolutions do. We first ask whether the learned routing actually exhibits the structure this design intends. To this end, we analyze the temporal dynamics of the channel-wise aggregation through a Temporal Routing Divergence (TRD), denoted $\mathcal{D}_{TR}$. For a given channel $c$, we compute its global temporal anchor $\bar{\mathbf{S}}_c$ as the time-averaged aggregation weight distribution across all $T$ frames, and define $\mathcal{D}_{TR}(c)$ as the expected Kullback-Leibler (KL) divergence between the instantaneous distribution $\mathbf{S}_c(t)$ and this anchor:
\begin{equation}
    \mathcal{D}_{TR}(c) = \frac{1}{T} \sum_{t=1}^{T} D_{\text{KL}}\!\left( \mathbf{S}_c(t) \,\|\, \bar{\mathbf{S}}_c \right).
    \label{eq:trd}
\end{equation}
This metric separates the channels into two regimes. Channels with a high $\mathcal{D}_{TR}$ actively modulate their reference selection across frames to blend temporal features dynamically, whereas channels with a low $\mathcal{D}_{TR}$ firmly anchor specific references and converge toward a stable weight distribution over time.

To test whether these two regimes carry distinct functional roles, we sort the channels by their $\mathcal{D}_{TR}$ values and perform knockout experiments, setting the ablated channels of $\mathbf{F}_{mix}$ to zero. As shown in Fig.~\ref{figs:TRD}, the two ablations produce strikingly orthogonal effects. Removing 50\% of the high-$\mathcal{D}_{TR}$ channels leaves rigid high-frequency details such as teeth perfectly sharp, but the flexible skin regions suffer severe spatial noise and rugged artifacts. Removing 50\% of the low-$\mathcal{D}_{TR}$ channels has the opposite effect: the facial skin becomes overly smooth while rigid details such as teeth melt into a blurred, averaged state. These observations validate a bipartite division of labor within the DTM. For low-frequency content such as continuous skin deformation, the high-$\mathcal{D}_{TR}$ channels dynamically modulate their weights to provide the adaptive mixing that static convolutions cannot achieve; for high-frequency content such as teeth, the low-$\mathcal{D}_{TR}$ channels supply a constant, stable reference that prevents homogenization. The DTM is therefore not an unconstrained router but one that adaptively anchors essential high-frequency features while remaining flexible elsewhere.

% \begin{figure}[t]
%     \centering
%     \includegraphics[width=0.45\textwidth]{figs/knockout.pdf}
%     \caption{\textbf{Causal knockout experiments guided by Temporal Routing Divergence ($\mathcal{D}_{TR}$).} $\mathcal{D}_{TR}$-guided channel ablations reveal how the DTM disentangles rigid high-frequency details from flexible low-frequency transitions.}
%     \label{figs:TRD}
% \end{figure}

\subsection{Temporal Dynamics of Channel Weights}
The routing analysis characterizes individual channels; we next study how the DTM allocates weight across references over time, and how this allocation depends on the reference pool. We define $\mathbf{S}^{n}_{\text{mean}}(t)$ as the channel-averaged routing weight assigned to the $n$-th reference at frame $t$, which summarizes its overall attention trajectory:
\begin{equation}
    \mathbf{S}^{n}_{\text{mean}}(t) = \frac{1}{C} \sum_{c=1}^{C} \mathbf{S}^{n}_{c}(t),
    \label{eq:smean}
\end{equation}
where $C$ is the total number of channels. We visualize these trajectories during cross-audio inference, comparing STAR Sampling against random sampling under varying reference budgets $N \in \{2, 3, 4, 5\}$. As Fig.~\ref{figs:comparison_routing} shows, under STAR Sampling the trajectories of different references fluctuate only mildly within distinctly separated, stable tiers, whereas under random sampling they intertwine chaotically and swing over a wide range.

To quantify this stability, we introduce the Intra-trajectory Temporal Volatility (ITV), the reference-averaged temporal standard deviation of these attention curves:
\begin{equation}
    \text{ITV} = \frac{1}{N} \sum_{n=1}^{N} \sqrt{ \frac{1}{T} \sum_{t=1}^{T} \left( \mathbf{S}^{n}_{\text{mean}}(t) - \mu^{n} \right)^{2} },
    \label{eq:itv}
\end{equation}
where $T$ is the number of frames and $\mu^{n}$ is the temporal mean of $\mathbf{S}^{n}_{\text{mean}}(t)$. A larger ITV indicates that the model's attention among the references shifts more violently over time. The annotated values in Fig.~\ref{figs:comparison_routing} reveal two consistent trends: ITV decreases as the number of references grows, and STAR Sampling maintains a strictly lower ITV than random sampling across every configuration. When the reference pool offers sufficient topological diversity, as under STAR Sampling, each anchor can stably retain its proportional contribution; the disordered weight shifting induced by random sampling instead disrupts temporal consistency, which plausibly underlies the synchronization degradation observed earlier at the feature level.

\subsection{Synergy between the DTM and STAR Sampling}
Taken together, the spatial routing and temporal volatility analyses expose a natural synergy between the two designs. The knockout results show that the DTM prefers a stable selection strategy to preserve high-frequency detail, and STAR Sampling provides exactly the conditions under which this strategy can operate. By maximizing the topological distance among selected references, STAR Sampling yields an unambiguous candidate pool, making it easy for the DTM to identify and lock onto the optimal high-frequency textures. This clarity manifests directly as the low ITV and the well-separated trajectory tiers seen under STAR Sampling in Fig.~\ref{figs:comparison_routing}. Random sampling, in contrast, returns topologically redundant or suboptimal references; faced with such candidates, the DTM cannot sustain its stable selection and is driven into the chaotic frame-to-frame weight shifting reflected by a high ITV, which ultimately compromises clarity and synchronization. STAR Sampling thus supplies the structural stability that lets the routing mechanism of the DTM reach its full potential.

\section{Conclusion}\label{sec:conclusion}
In this paper, we present EfficientSync, a novel deformation-based lip-sync framework designed to generate high-fidelity dubbed videos while maintaining a low memory footprint, low computational complexity, and high inference speeds. To preserve authentic facial details and avoid texture degradation, we introduce the Dynamic Texture Mixer for context-aware reference feature aggregation. Furthermore, we propose the Spatio-Temporal Shifted Adaptive Masking strategy to effectively resolve the background preservation dilemma frequently overlooked by previous mask-based methods, thereby eliminating boundary artifacts. Additionally, we introduce the STAR Sampling strategy, which optimizes reference image selection to enhance final generation quality without incurring any additional inference latency. By achieving robust identity preservation and a real-time inference speed of 166 FPS, we hope these improvements will facilitate the broader real-world deployment of computationally efficient lip-sync models.
\section*{Acknowledgement}
This work is supported in by the Guangdong Basic and Applied Basic Research Foundation (2025A1515010124), and in part by the National Key Research and Development Program of China (2024YFF1206600), the National Natural Science Foundation of China(62325204).

% or


% \appendices

% \section{}

\ifCLASSOPTIONcompsoc
%   \section*{Acknowledgments}
% \else
%   % regular IEEE prefers the singular form
% \fi
% This research is supported in part by
% the Early Career Scheme of the Research Grants Council
% (RGC) of the Hong Kong SAR under grant No. 26202321,
% HKUST Startup Fund No.~R9253.

\ifCLASSOPTIONcaptionsoff
  \newpage
\fi



%\bibliographystyle{IEEEtran}
\bibliographystyle{ieeetr}
\bibliography{tpami}


% \end{thebibliography}
\begin{IEEEbiography}[{\includegraphics[width=1in,height=1.25in,clip,keepaspectratio]{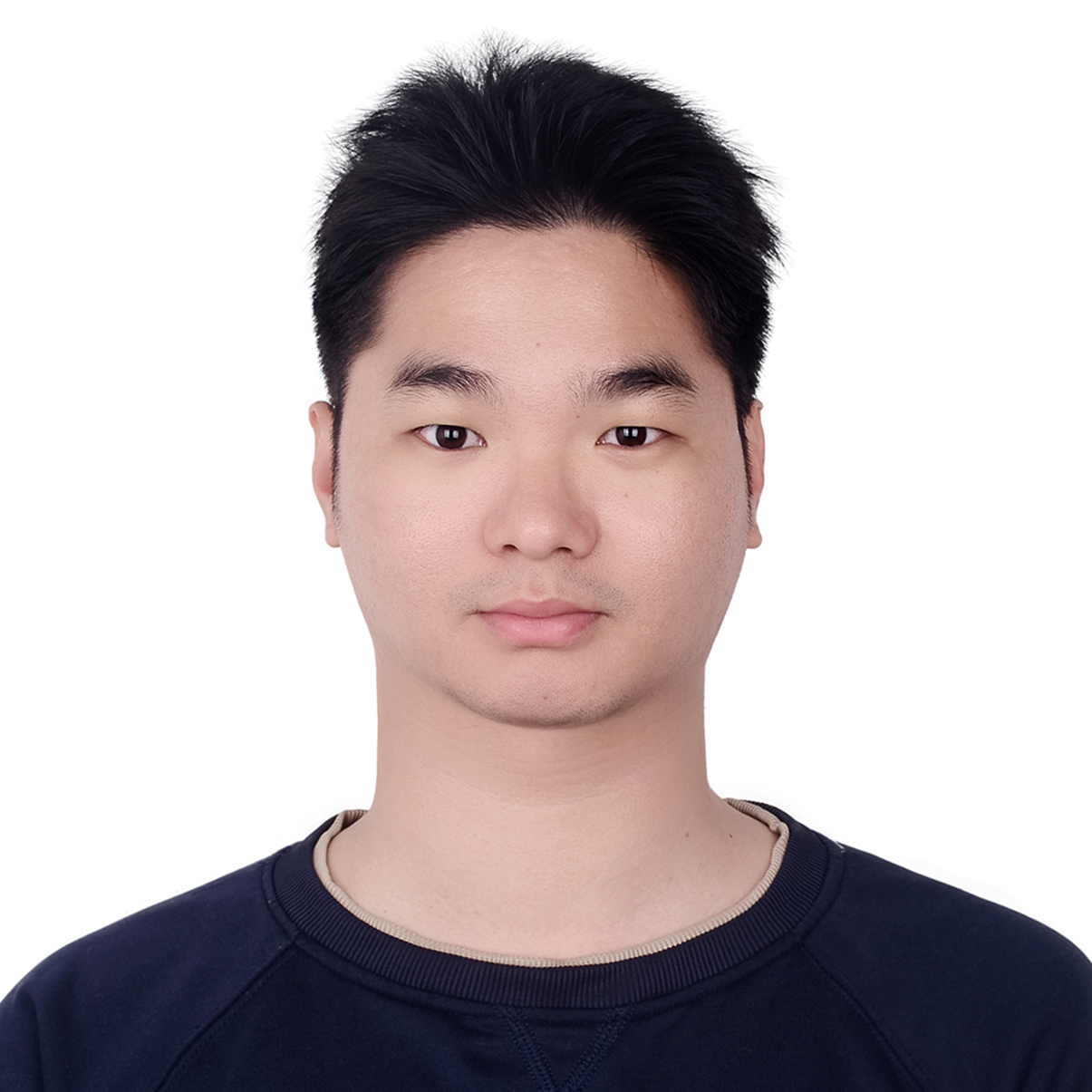}}]{Fa-Ting Hong} did his PhD at Hong Kong University of Science and Technology. He received his M.E. from Sun Yat-sen University and B.E. from South China University of Technology. His research interest lies in deep learning for avatar generation, human motion and audio-visual learning.
\end{IEEEbiography}
\vspace{-10pt}

\begin{IEEEbiography}[{\includegraphics[width=1in,height=1.25in,clip,keepaspectratio]{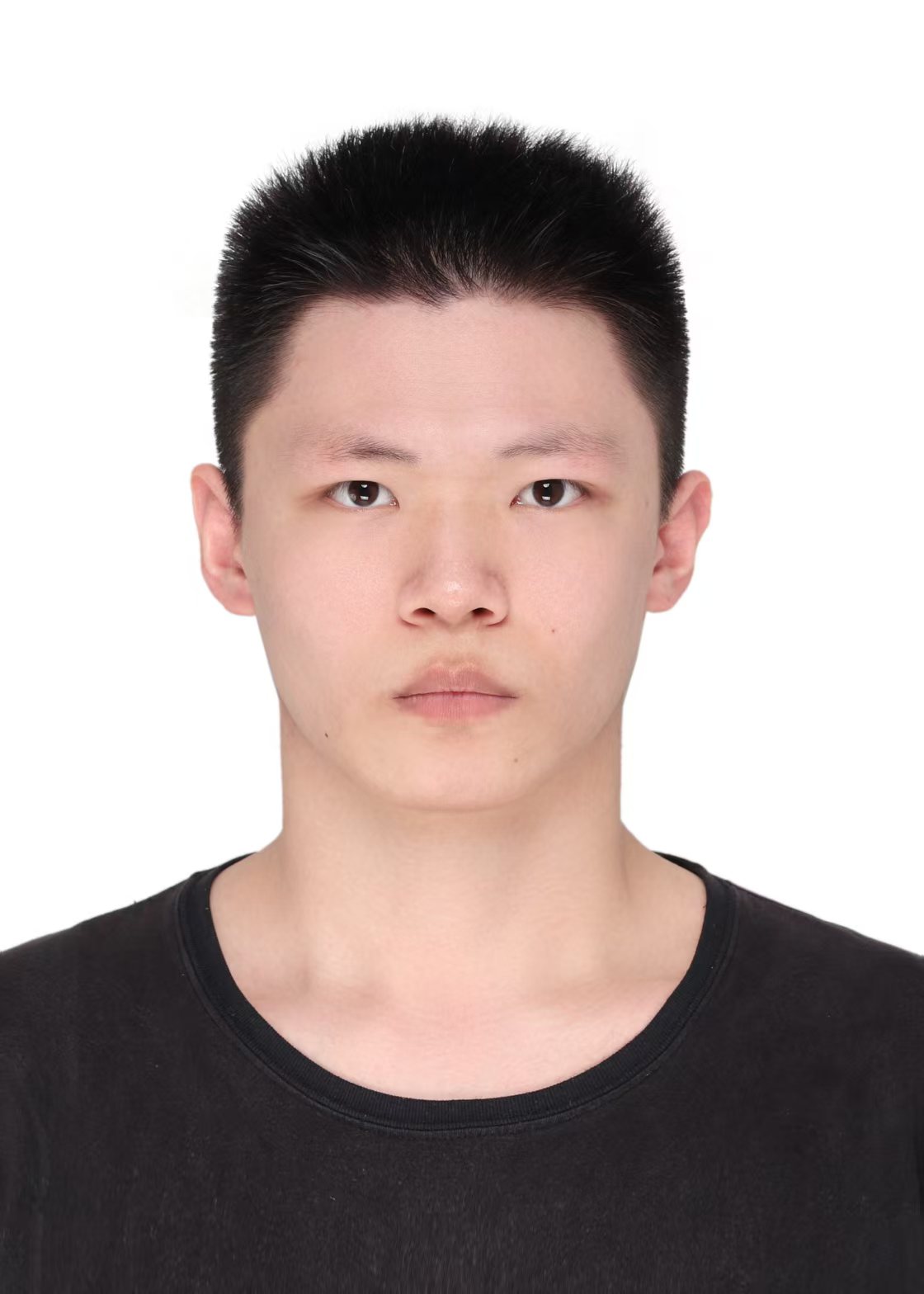}}]{Runzhen Liu} is a master's student in the School of Computer Science and Engineering, South China University of Technology. He received his B.E. from South China University of Technology. His research interest lies in deep learning for image and video generation.
\end{IEEEbiography}
\vspace{-10pt}

\begin{IEEEbiography}[{\includegraphics[width=1in,height=1.25in,clip,keepaspectratio]{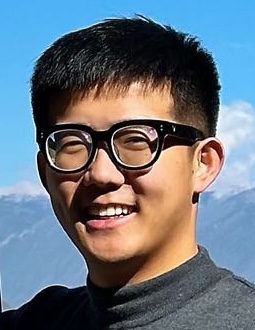}}]{Luchuan Song} is a Ph.D. student in the Department of Computer Science at University of Rochester. He received his M.E. and B.E. from Department of Electronic Engineering and Information Science at University of Science and Technology of China. His research interest lies in human related topic (\textit{e.g.} face animation and toonfication, 3D face reconstruction, deepfake detection \textit{e.t.c}).
\end{IEEEbiography}
\vspace{-10pt}

\begin{IEEEbiography}[{\includegraphics[width=1in,height=1.25in,clip,keepaspectratio]{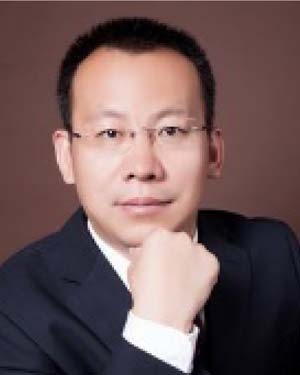}}]{Hongmin Cai}  (Senior Member, IEEE) received the B.S. and M.S. degrees in mathematics from the Harbin Institute of Technology, Harbin, China, in 2001 and 2003, respectively, and the Ph.D. degree in applied mathematics from the University of Hong Kong, Hong Kong, in 2007. He was a visiting Professor with Kyoto University, Kyoto, Japan, and with Harvard University, Cambridge, MA, USA. He is currently the Executive Dean with the School of Future Technology, and a Joint Appointment Professor with the School of Computer Science and Engineering, South China University of Technology, Guangzhou, China. His research interests include digital twins in life science, intelligent computing of biomedical data, and machine learning.
\end{IEEEbiography}
\vspace{-10pt}

\begin{IEEEbiography}[{\includegraphics[width=1in,height=1.25in,clip,keepaspectratio]{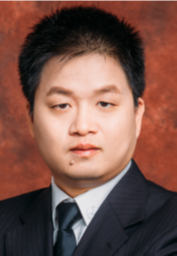}}]{Chuhua Xian} is an Associate Professor with the School of Computer Science and Engineering, South China University of Technology. He received Ph.D. degree in Computer Science and Technology from State Key Lab. of CAD \& CG of Zhejiang University in March 2012. From November 2013 to May 2014, and from September 2015 to April 2016, he was a visiting scholar of the research group of Prof. Charlie C. L. Wang in the Chinese University of Hong Kong. His current research interests are computer graphics, geometry computing, rendering, image processing, and Computer-Aided Design.
\end{IEEEbiography}

% \clearpage
% \input{context/coverletter.tex}

%\newpage



%\vfill


\end{document}